\pdfoutput=1

\documentclass[11pt]{article}

\usepackage{acl}

\usepackage{times}
\usepackage{latexsym}

\usepackage[T1]{fontenc}

\usepackage[utf8]{inputenc}

\usepackage{microtype}
\usepackage{graphicx}
\usepackage{url}
\usepackage{amsfonts}
\usepackage{amsmath}
\usepackage{amssymb}
\usepackage{physics}
\usepackage{xcolor}
\usepackage{subfiles}
\usepackage[mathscr]{euscript}
\usepackage{bbm}
\usepackage{bm}
\usepackage{caption}
\usepackage{subcaption}

\title{Capturing Perspectives of Crowdsourced Annotators in Subjective Learning Tasks}

\author{Negar Mokhberian$^1$, Myrl G. Marmarelis$^1$, Frederic R. Hopp$^2$,\\ {\bf Valerio Basile$^3$, Fred Morstatter$^1$, Kristina Lerman$^1$} \\
        \{nmokhber, myrlm, fredmors, lerman\}@isi.edu, f.r.hopp@uva.nl, valerio.basile@unito.it \\ $^1$Information Sciences Institute, University of Southern California \\ 
        $^2$School of Communication Research, University of Amsterdam \\ 
        $^3$Computer Science Department, University of Turin}

\begin{document}
\maketitle
\begin{abstract}
Supervised classification heavily depends on datasets annotated by humans. However, in subjective tasks such as toxicity classification, these annotations often exhibit low agreement among raters. Annotations have commonly been aggregated by employing methods like majority voting to determine a single ground truth label. In subjective tasks, aggregating labels will result in biased labeling and, consequently, biased models that can overlook minority opinions. Previous studies have shed light on the pitfalls of label aggregation and have introduced a handful of practical approaches to tackle this issue. Recently proposed multi-annotator models, which predict labels individually per annotator, are vulnerable to under-determination for annotators with few samples. This problem is exacerbated in crowdsourced datasets. In this work, we propose \textbf{Annotator Aware Representations for Texts (AART)} for subjective classification tasks. Our approach involves learning representations of annotators, allowing for exploration of annotation behaviors. 
We show the improvement of our method on metrics that assess the performance on capturing individual annotators' perspectives. Additionally, we demonstrate fairness metrics to evaluate our model's equability of performance for marginalized annotators compared to others.\footnote{Our code and implementation are publicly available at \href{https://github.com/negar-mokhberian/aart}{https://github.com/negar-mokhberian/aart}.}
\end{abstract}

\begin{figure*}[ht!]
\centering
\includegraphics[width=\textwidth]{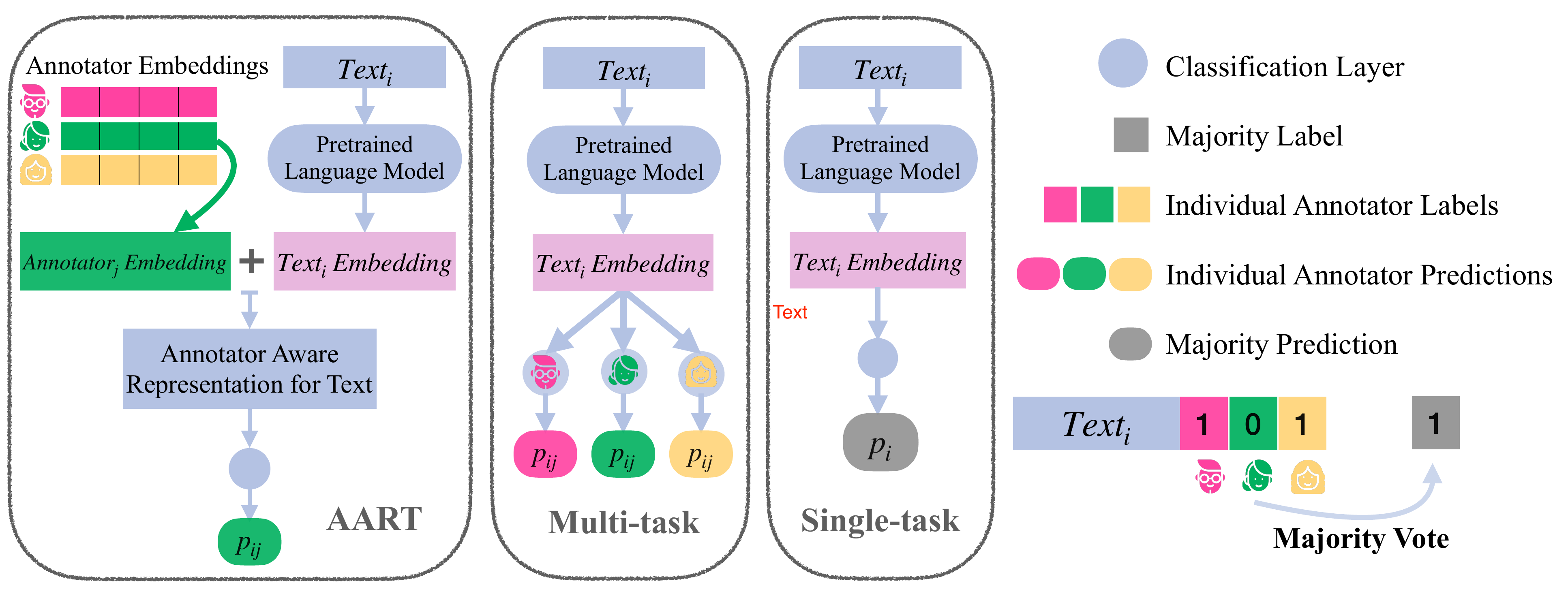}
\caption{The architecture of AART compared to the multi-task and single-task approaches. Whereas a multi-task model incorporates annotator subjectivity by training entirely separate classification heads per annotator, AART represents annotator-specific biases as embeddings that are added to the text embeddings prior to classification. This new approach constrains the degrees of freedom and allows a common head across annotators, lending helpful inductive biases to sparse annotators.}
\label{fig:model}
\end{figure*}
\section{Introduction}
Supervised machine learning tasks often train on data provided by human annotators. To improve the quality of ground truth data, the data samples are labeled by multiple annotators, whose independent decisions are then aggregated, often by taking the majority label or average value. This approach works well in some tasks, like image recognition, where ``objective'' ground truth exists. Objective judgments should not be affected by individual differences among annotators and in practice, most people would perceive and classify them similarly, except for obvious errors or oversight~\cite{basile2021toward}.

However, some annotation tasks are subjective by nature, and traditional approaches to aggregating annotations that ignore context and subjectivity may lose important information~\cite{plank-2022-problem}. For instance, individuals may differ about what they consider to be toxic or hate speech on social media, depending on their background and attitudes~\cite{sap-etal-2022-annotators}. As a result, not all annotators will agree when a particular message constitutes hate speech. Disagreement among annotators in subjective tasks is expected because, for each item, the annotation outcome depends on (1) the annotator's perspective and lived experience, (2) the item being evaluated, (3) the time and circumstances surrounding the annotation task, and (4) the clarity of the designed prompts or instructions given to annotators~\cite{basile-etal-2021-need, denton2021whose}. In addition, annotators vary in how much effort they put into the task, with prolific annotators labeling many samples and sparse annotators labeling few samples. A classifier that does not take variety in human labels into account risks overlooking minority perspectives by ignoring potentially informative differences between annotators and integrating the biases of a particular annotator pool.  

Historically, most public datasets did not provide labels at the individual annotator level. This limited the ability of machine learning researchers to develop methods that take advantage of differences in annotator perspectives in subjective learning tasks. Recently, researchers have acknowledged the importance of incorporating this information and have released raw annotations, often accompanied by demographic information about the annotators. It has now become commonplace to include multiple annotations per item in published datasets~\cite{frenda-etal-2023-epic, sap-etal-2022-annotators, prabhakaran-etal-2021-releasing, denton2021whose}. Recent studies have revealed that the aggregation of annotations exhibits detrimental biases toward specific groups of annotators \cite{prabhakaran-etal-2021-releasing, aroyo2024dices}.

Instead of adhering to the conventional notion that labels should be aggregated across all stages of AI pipelines, recent research advocates for leveraging the inherent diversity in human-assigned labels by pioneering innovative approaches to accommodate multiple votes per item at different stages of AI pipelines. This can be applied across domains such as model architecture design \cite{weerasooriya-etal-2023-disagreement, davani-etal-2022-dealing, hayat_sensors, gordon2022jury, deng-etal-2023-annotate, gordon2021disagreement}, performance evaluation metrics \cite{basile-etal-2021-need, dumitrache-etal-2019-crowdsourced, gordon2021disagreement}, or label aggregation, surpassing simplistic methods like averaging or majority voting \cite{aroyo2014three, dumitrache2018crowdtruth}.

Despite some progress, few researchers have systematically  tackled the problem of learning from multiple annotations. We address this gap by proposing a new approach for modeling annotator perspectives in subjective classification tasks via \textbf{Annotator Aware Representations for Texts (AART)}. Our approach learns annotator embeddings that represent their perspectives in downstream classification tasks (see \S\ref{sec:method} for more details). We demonstrate empirically that AART affords a more accurate and context-sensitive interpretation of subjective data at the individual annotator level.

We compare our approach to a widely recognized method~\cite{davani-etal-2022-dealing} that captures differences in annotator judgments using a multi-task architecture consisting of a separate trainable classification-head per annotator. We show that the multi-task learning suffers from having under-trained classification heads for \textit{sparse} annotators that contribute few annotations. Although this work marks an advance in the study of annotations of subjective tasks, its difficulties in accounting for differences in perspectives of infrequent, sparse annotators limits its utility in practical applications. 

In contrast, our method captures the semantics inherent in subjective classification tasks, accounting for the diverse perspectives of annotators, including those who contribute few annotations. The AART architecture combines text embedding with annotator embeddings and provides this joint information to the classification layer. We couple this design choice with using the L2 regularizer in the loss, to prevent annotator embeddings from overfitting. Additionally, we use the contrastive loss to aid in learning from the similarities and dissimilarities of label choices among annotators, a capability lacking in Multi-task. Furthermore, in Section \ref{sec:experiments}, we examine the learned annotation behaviors.

By leveraging metrics from algorithmic fairness, we demonstrate that the viewpoints of sparse annotators---who either contribute few samples or select less popular labels---are accurately represented (See \S \ref{sec:fairness}). This inclusivity is crucial for applications where the diversity of perspectives is essential for understanding patterns in subjective annotations. Practitioners can utilize this approach to anticipate how individuals and, by extension, subgroups would react to a particular input.



\section{Related Work}
\paragraph{Sources of Annotator Disagreement.}

Some previous research has shown aggregating the labels takes away the voice of minority socio-demographic groups \cite{prabhakaran-etal-2021-releasing}. \citet{al-kuwatly-etal-2020-identifying} conducted a study grouping annotators according to socio-demographic attributes and discovered biases in the annotations of personal attack corpora is linked to annotators' age, education, and first language. Similarly, \cite{sap-etal-2022-annotators} found that posts written in African American dialect were more likely to be labeled as offensive by conservative annotators. Furthermore, \cite{larimore-etal-2021-reconsidering} demonstrated that there are differing interpretations among white and non-white workers when annotating racially sensitive topics, indicating the importance of considering the perspectives of those affected when gathering annotations of racist language.

It is important to note that many public datasets lack information regarding the demographics of the annotators alongside their annotations. Moreover, even in cases where datasets do report annotator demographics, mostly there exists an imbalance in the representation of demographics among annotators~\cite{orlikowski-etal-2023-ecological}. For instance, while females constitute the majority of crowd workers in the US, they are significantly underrepresented in most other countries. Additionally, online platforms tend to attract a predominantly young demographic of annotators \cite{Posch_2022, hara2019worker}. 

However, it is essential to recognize that relevant lived experience and mindset among annotators do not always perfectly align with demographic characteristics \cite{orlikowski-etal-2023-ecological}. \citet{waseem2016you} found that amateur annotators were more prone to mislabel content as racist or sexist compared to experts with backgrounds in feminism and anti-racism activism. This finding emphasizes the need for researchers to identify annotator behaviors beyond relying solely on demographic characteristics. Hence, some previous work has explored methods to identify groups of annotators with similar annotation behaviors. \citet{wich-etal-2020-investigating} employed graph methods, and \cite{DBLP:conf/nlperspectives/LoB23} conducted hierarchical clustering of annotators, while \cite{akhtar2020modeling} and \cite{basile2021s} utilized exhaustive-search algorithms to automatically group human annotators in order to maximize the inter-group differences in annotation patterns.


\paragraph{Modeling Annotator Disagreement.}


Examining annotation disagreement has been a subject of interest in previous research. One line of research treats disagreement as noise and attempts to minimize it in order to improve performance \cite{mokhberian2022noise, hovy-etal-2013-learning}. However, another line of research argues in favor of embracing annotation disagreement, considering it as a valuable signal rather than entirely noise \cite{Aroyo_Welty_2015, learncrowd}. Our paper explores the latter possibility.

 Certain studies have demonstrated that the level of disagreement among annotators for each data point can be predicted based on the text itself \cite{kenyon-dean-etal-2018-sentiment}. Additionally, incorporating demographic information of the annotators has shown even better performance~\cite{wan2023everyone}. 

Recent work for numerous tasks has proposed to approximate the full annotators' label distribution on each item, instead of aggregating, utilizing methods such as soft-label learning \cite{fornaciari-etal-2021-beyond, uma-etal-2021-semeval, uma2022scaling}. 

Other research has developed approaches to model annotators for subjective tasks and to predict a personalized label for each annotator. \citet{plepi2022unifying} explore methods of personalization to model Reddit users and analyze the effect of their method on demographic groups.
\citet{akhtar2020modeling} employ a comprehensive search approach to divide annotators into two groups based on their annotation patterns. They demonstrate the effectiveness of an ensemble model comprising two separate classifiers representing the perspectives of each group over the traditional single-task model, which merely observes aggregated labels. \citet{davani-etal-2022-dealing} explored the effectively of three different multi-annotator architectures for predicting each annotator's labels. \citet{deng-etal-2023-annotate} integrate annotator embeddings into their model design, yet they were unable to identify significant annotator patterns solely based on these embeddings without incorporating additional embeddings for representing annotation labels. \citet{gordon2022jury} introduced the concept of jury learning, a recommender system approach that specifies the demographic composition of individuals, along with their respective proportions, tasked with determining the predictions of the classifier. For example, in the context of online hate speech detection, a jury learning model might recommend the inclusion of women and Black jurors, as they are often primary targets of online harassment. Drawing motivation from this line of work, our paper investigates a new approach to model the subjectivity of tasks through label variations among individual annotators. By training annotator-specific embeddings with a contrastive loss, we learn a latent space that captures similarities and differences in annotator behavior.

\section{Methodology}
\label{sec:method}
This section presents the details of our proposed approach and baselines used for evaluation. Our proposed method \textit{\textbf{AART}} (\S\ref{sec:aart}) learns separate representations for each annotator in an end-to-end fashion during the training phase of the main classification task. In the inference phase, we are able to predict personalized labels for each annotator based on the annotation behaviors observed during training. A commonly used method for learning subjective perspectives is \textit{\textbf{multi-task}} model (\S\ref{sec:multi_task}) which trains a separate classification head for each of the annotators. Other than comparing our method to the multi-task model we also compare it to a \textit{\textbf{single-task}} (\S\ref{sec:single_task}) model which is fed the aggregated labels during training and is not aware of the variety of annotations for each item. The overall framework of these approaches is shown in Fig.~\ref{fig:model}.

\subsection{Problem Definition}
\label{sec:problem_def}
We consider an input corpus $\mathcal{D}=\{(x_i, a_j, y_{ij})\}$ that is composed of triplets, with elements coming from text items $X=\{x_i\}_{i=1}^N$, annotators $A=\{a_j\}_{j=1}^M$, and annotations $ y_{ij}\in\{1, ..., Q\}$. Each $(i,j)$ combination that appears (at most once) in $\mathcal{D}$ means that annotator $a_j$ assigned label $y_{ij}$ to the text item $x_i$. 
In most annotated datasets, many $y_{ij}$ values are not available as annotators only label a portion of items. 

 Most traditional classification studies aim to predict the aggregated label that would be assigned to text $x_i$ by the annotator pool $\{a_j\}_{j=1}^M$. This aggregated label is usually calculated via majority-voting or averaging. This fundamental model, referred to as the \textit{single-task} model in this paper (See \S\ref{sec:single_task}), solely considers and predicts the aggregated labels.
In this study, however, we are interested in predicting the label that annotator $a_j$ would assign to an unseen text item $x_i$, i.e., predict $y_{ij}$ given a pair of annotator and item $(x_i, a_j)$. We use the overarching term ``multi-annotator models" to encompass these models, and we will describe our proposed multi-annotator model, \textit{AART} (see \S \ref{sec:aart}), and the established \textit{multi-task} (see \S \ref{sec:multi_task}) model found in prior literature.

All the approaches we investigate in this paper leverage pretrained transformer-based language models for encoding text (such as BERT \cite{devlin-etal-2019-bert} and RoBERTa \cite{liu2019roberta}). Given a text item $x_i$, we extract its generic representation by taking the embedding of the [CLS] token from the final layer of the language model, denoted as $e(x_i)\in \mathbb{R}^{d}$ ($d$ is the size of the representation vector defined in the language model). This representation is then fine-tuned in conjunction with the other parameters of the model architectures during training.

\subsection{Annotator-Aware Representations for Texts (AART)}
\label{sec:aart}
We present a novel approach for incorporating annotator information into text representations for the purpose of learning an annotator's label on a given text instance. Our approach utilizes $M$ annotator embeddings, with one randomly initialized embedding per annotator, as learnable parameters with the same dimensionality as the text embeddings. Given a text instance $x_i$ and annotator $a_j$, the annotator-aware text embedding $g(x_i, a_j)$ is calculated as the sum of the text embedding $e(x_i)$ and annotator embedding $f(a_j)$, yielding $g(x_i, a_j)= e(x_i) + f(a_j)$. We also tried concatenating a one-hot encoding for each annotator with text embeddings which resulted in lower performance compared to taking the sum.
Our idea draws inspiration from the positional encoding used by transformers \cite{vaswani2017attention}, where a positional embedding of the same dimensionality as the token embedding is added to the token embedding at each position to form the sequence representation. A similar approach has been found effective in dialogue act classifications by learning speaker-turn embeddings in a conversation \cite{he-etal-2021-speaker-turn}. We feed the annotator-aware representations of text $g(x_i, a_j)$ into a fully connected classification layer.

To train AART we propose the following objective function:
    $$\mathcal{L} = \mathcal{L}_{CE} + \lambda \sum_{j} \norm{f(a_j)}_2 + \alpha\sum_{j,j'}\mathcal{L}_C(j,j') $$

Where the first term is a cross-entropy loss for predicting the item-annotator labels $y_{ij}$, the second term is the regularizing $L_2$ norm for annotator embeddings, and the third term is contrastive loss. We calculate InfoNCE contrastive loss \cite{oord2018representation, chen2020simple} on pairs of $a_{j}$ and $a_{j'}$ that have both worked on the same text $x_i$. When two annotators $a_j$ and $a_{j'}$ have assigned the same label to $x_i$ we consider $a_{j}$ and $a_{j'}$ as a positive pair and our loss will push them closer in the latent space. We consider a negative pair to be any two annotators that have assigned different labels to that item. The contrastive loss for a positive pair of annotators $(j, j')$ is defined as below: 

\begin{multline*}
    \mathcal{L}_{C}(j, j') = \\-\log \frac{\exp{-\norm{a_{j}-a_{j'}}^2_2/\tau}}{\sum_{k\in\mathcal{N}_j \cup \{j'\}} \exp{-\norm{a_{j}-a_{k}}^2_2/\tau}},
\end{multline*}


in which $\tau$ is a temperature constant and $\mathcal{N}_j$ is the set of all negatives for annotator $j$. 

\subsection{Multi-task Approach}
\label{sec:multi_task}
The multi-task approach learns each annotator's labeling behavior via individual classification tasks, utilizing shared language model layers to derive a consistent representation of the input text, $e(x_i)$, for all annotators~\cite{davani-etal-2022-dealing}. 
To capture annotator-specific labeling behavior, this method has a separate fully connected layer specifically optimized for each annotator.
However, the text encoding layers generating $e(x_i)$ are refined based on the outputs of all annotator tasks. 
The optimization objectives are formulated separately for each annotator as a cross-entropy loss function, which takes into account each annotator's available labels for each item $x_i$. 

\subsection{Single-task Approach}
\label{sec:single_task}
The baseline model embodies the typical approach of using a single-task classifier to predict the aggregated label for each instance. This model is constructed by adding a fully connected classification layer to the text embedding $e(x_i)$. The fully connected layer performs a linear transformation in order to predict the majority label. Unlike the multi-annotator models detailed in \S\ref{sec:aart} and \S\ref{sec:multi_task}, the baseline model does not incorporate the individual annotator labels $y_{ij}$; rather, it directly predicts the aggregated label $\bar y_i$ for each item $x_i$.

\begin{table*}[ht!]
\centering
\small
\begin{tabular}{lcccc}
\multicolumn{1}{l|}{\textit{Dataset}}       & \textit{Single-task }     & \textit{Multi-task}  & \textit{AART} ($\alpha=0$) & \textit{AART} ($\alpha>0$)   \\[0.7mm] \hline
\multicolumn{5}{c}{\textit{Annotator-level F1}}         \\ \hline
\multicolumn{1}{l|}{$\mathcal{D}_{\textsl{MDA}}$} & $66.80 \pm 0.7$ & $61.87 \pm 1.5$ & $\underline{69.64 \pm 1.3}$ & \bm{$69.72 \pm 1.1$} \\[0.7mm]
\multicolumn{1}{l|}{$\mathcal{D}_{\textsl{I}}$}          & $\underline{58.59 \pm 1.9}$ & $57.31 \pm 0.9$ & $56.45 \pm 7.9$ & \bm{$59.67 \pm 0.9$} \\[0.7mm]
\multicolumn{1}{l|}{$\mathcal{D}_{\textsl{RB}}$}          & $68.61 \pm 1.5$ & $66.57 \pm 3.1$ & \bm{$71.6 \pm 1.6$}  & $\underline{71.1 \pm 3.2}$  \\[0.7mm]
\hline
\multicolumn{5}{c}{\textit{Global-level F1}}       
\\ \hline
\multicolumn{1}{l|}{$\mathcal{D}_{\textsl{MDA}}$} & $71.99 \pm 0.6$ & $74.19 \pm 0.9$ & \bm{$77.40 \pm 0.6$} & $\underline{77.38 \pm 0.4}$ \\[0.7mm]
\multicolumn{1}{l|}{$\mathcal{D}_{\textsl{I}}$}          & $60.23 \pm 1.7$ & \bm{$66.78 \pm 0.9$} & $63.64 \pm 5.4$ & $\underline{66.16 \pm 1.4}$ \\[0.7mm]
\multicolumn{1}{l|}{$\mathcal{D}_{\textsl{RB}}$}          & $71.97 \pm 1.7$ & $77.72 \pm 1.8$ & $\underline{79.38 \pm 2.4}$ & \bm{$79.96 \pm 1.9$} \\[0.7mm]
\hline
\multicolumn{5}{c}{\textit{Item-level Disagreement Correlations}}        
\\ \hline
\multicolumn{1}{l|}{$\mathcal{D}_{\textsl{MDA}}$} &               \small{NA}               & $0.33 \pm 0.03$ & \bm{$0.39 \pm 0.03$} & $\underline{0.37 \pm 0.04}$ \\[0.7mm]
\multicolumn{1}{l|}{$\mathcal{D}_{\textsl{I}}$}          &           \small{NA}                   & \bm{$0.23 \pm 0.04$} & $0.14 \pm 0.10$ & $\underline{0.20 \pm 0.06}$ \\[0.7mm]
\multicolumn{1}{l|}{$\mathcal{D}_{\textsl{RB}}$}          &           \small{NA}                   & $0.47 \pm 0.05$ & \underline{$0.48 \pm 0.12$} & \bm{$0.54 \pm 0.04$}
\\[0.7mm] \hline
\end{tabular}
\caption{This table presents experimental results for the classification tasks. AART is the new method introduced here. Single-task is a baseline that only sees and predicts aggregated labels. We repeated each experiment 10 times and report the mean and standard deviation. $\alpha$ is the coefficient for contrastive loss. The best performing model is indicated in bold and the second best is underlined. See section \S\ref{sec:experiments} for further details.}
\label{tab:results}
\vspace{-1em}
\end{table*}

\section{Experiments}
\label{sec:experiments}

\subsection{Datasets}

\begingroup
\setlength{\tabcolsep}{2pt} 
\renewcommand{\arraystretch}{1} 

\begin{table}[ht!]
\centering
\caption{Statistics of the datasets used in our experiments. $\mathcal{D}_{\textsl{MDA}}$ refers to the offensive speech detection dataset~\cite{leonardelli-etal-2021-agreeing}, $\mathcal{D}_{\textsl{I}}$  refers to the irony detection corpus~\cite{frenda-etal-2023-epic}, and  $\mathcal{D}_{\textsl{RB}}$  to the racial bias corpus~\cite{sap-etal-2019-risk}.}
\label{tab:datasets}
\small
\begin{tabular}{llll}
                             & $\mathcal{D}_{\textsl{MDA}}$       & $\mathcal{D}_{\textsl{I}}$            & $\mathcal{D}_{\textsl{RB}}$       \\
                             \hline
\# Annotators                & 819                       & 74                           & 334                      \\
\# Annotations per Annotator & $60 \pm 131$ & $192 \pm 27$ & $35 \pm 51$ \\
\# Unique Texts              & 9814                      & 2884                         & 1309                     \\
\# Annotations per Text      & 5                         & 5                            & $9 \pm 2$  
\end{tabular}
\end{table}
\endgroup

\paragraph{The Multi-Domain Agreement dataset ($\mathcal{D}_{\textsl{MDA}}$):} \cite{leonardelli-etal-2021-agreeing} created an dataset for the task of offensive language detection. The dataset consists of 9,814 English tweets from three domains (Black Lives Matter movement, Election 2020, and COVID-19 pandemic). Each tweet was annotated for offensiveness by 5 annotators via Amazon Mechanical Turk. 

\paragraph{English Perspectivist Irony Corpus ($\mathcal{D}_{\textsl{I}}$):}
The irony detection corpus, as presented by \citet{frenda-etal-2023-epic}, encompasses 3,000 Post-Reply pairs extracted from social media. These pairs are evenly sourced from Twitter and Reddit, originating from five English-speaking countries: Australia, India, Ireland, the United Kingdom, and the United States. For Twitter data, the authors utilized the API geolocation service to distinguish the five English varieties. In the case of Reddit, data was gathered from specific subreddits, assuming the texts' origins: r/AskReddit (United States), r/CasualUK (United Kingdom), r/britishproblems (United Kingdom), r/australia (Australia), r/ireland (Ireland), r/india (India). The 74 annotators were balanced in terms of both gender and nationality, with approximately 15 raters assigned to each of the mentioned nationalities. Each rater labeled around 200 instances, resulting in a corpus containing 14,172 annotations, with a median of 5 annotations per instance.

\paragraph{The Racial Bias Toxicity Detection Corpus ($\mathcal{D}_{\textsl{RB}}$):}
\citet{sap-etal-2019-risk} have studied the effect of annotators' biases on their toxicity perceptions. They have confirmed in their work that determining what is harmful is a subjective task and annotators' racial biases can transmit to their annotations. They have gathered toxicity annotations for tweets on Amazon Mechanical Turk asking workers to determine whether a tweet (a) is offensive to them, and (b) could be seen as offensive to anyone. We work with the former as personal perspectives are more relevant to this study. 

\subsection{Evaluation Metrics}
\label{sec:eval}
\paragraph{Annotator-level F1}
This study is primarily motivated by the objective of safeguarding minority perspectives that might be overlooked when resolving annotator disagreements through the common practice of aggregating labels in machine learning. Consequently, it becomes crucial to demonstrate performance of model across all annotators, regardless of whether they have made substantial or minimal contributions to the annotated dataset. To address this, we introduce the metric of \textit{Annotator-level F1} defined as the average macro F1 score across all annotators. The performance for each annotator $a_j$ over items $x_i$ appearing in the test split is calculated by comparing the existing ground truths $y_{ij}$ to the model's predicted label for $(x_i, a_j)$.

\paragraph{Global-level F1}{The metric \textit{Global-level F1} considers all pairs of item-annotator $(x_i, a_j)$ appearing in the test set and calculates the macro F1 comparing their true label $y_{ij}$ to the predicted label of the model for $(x_i, a_j)$. In contrast to the Annotator-level F1, this metric implicitly gives preference to larger annotators, potentially overshadowing the performance of sparse annotators within the substantial contributions of prolific annotators.
}
\paragraph{Disagreement Measurement for Items}
In the context of multi-annotator models, a crucial consideration is whether the degree of disagreement in the predicted labels aligns with the actual degree of disagreement for each item. We follow the definitions from \cite{wan2023everyone} for quantifying disagreement for each item. They introduce a continuous disagreement metric, quantifying the fraction of votes for an item that differs from the majority vote. For instance, if five votes for an item are recorded as 0,1,0,1,1, the disagreement is computed as $0.4$. The disagreement scale ranges from 0 (complete unanimity in annotation results) to 1 (a significant divergence in opinions on annotation results). However, in the case of binary labels, the disagreement scale is confined to the range $[0, 0.5]$. Notably, we calculate predicted disagreement for each item exclusively based on annotators for whom ground truth information is available, aligning with the set of annotators used to calculate the ground truth disagreement per item.

\begin{figure*}[ht!]
     \centering
     \begin{subfigure}[b]{0.25\textwidth}
         \centering
         \includegraphics[width=\textwidth]{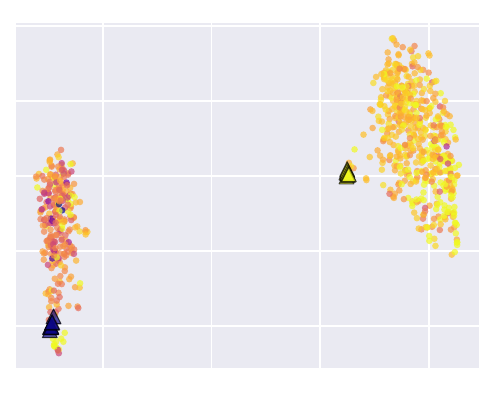}
         \caption{$\mathcal{D}_{\textsl{MDA}}$}
     \end{subfigure}
     \begin{subfigure}[b]{0.25\textwidth}
         \centering
         \includegraphics[width=\textwidth]{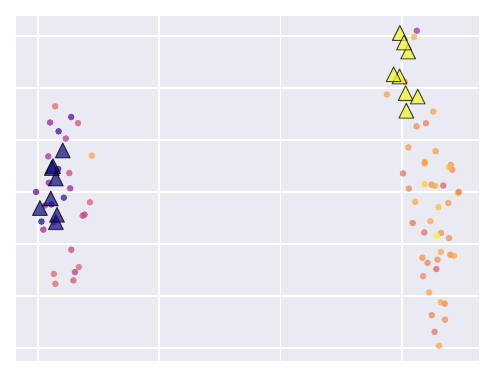}
         \caption{$\mathcal{D}_{\textsl{I}}$}
     \end{subfigure}
     \begin{subfigure}[b]{0.25\textwidth}
         \centering
         \includegraphics[width=\textwidth]{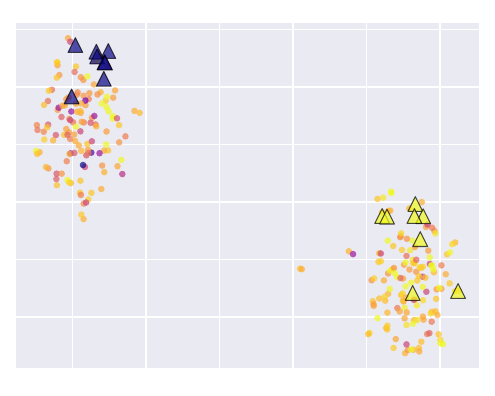}
         \caption{$\mathcal{D}_{\textsl{RB}}$}
     \end{subfigure}
     \begin{subfigure}[b]{0.15\textwidth}
         \centering
         \includegraphics[width=\textwidth]{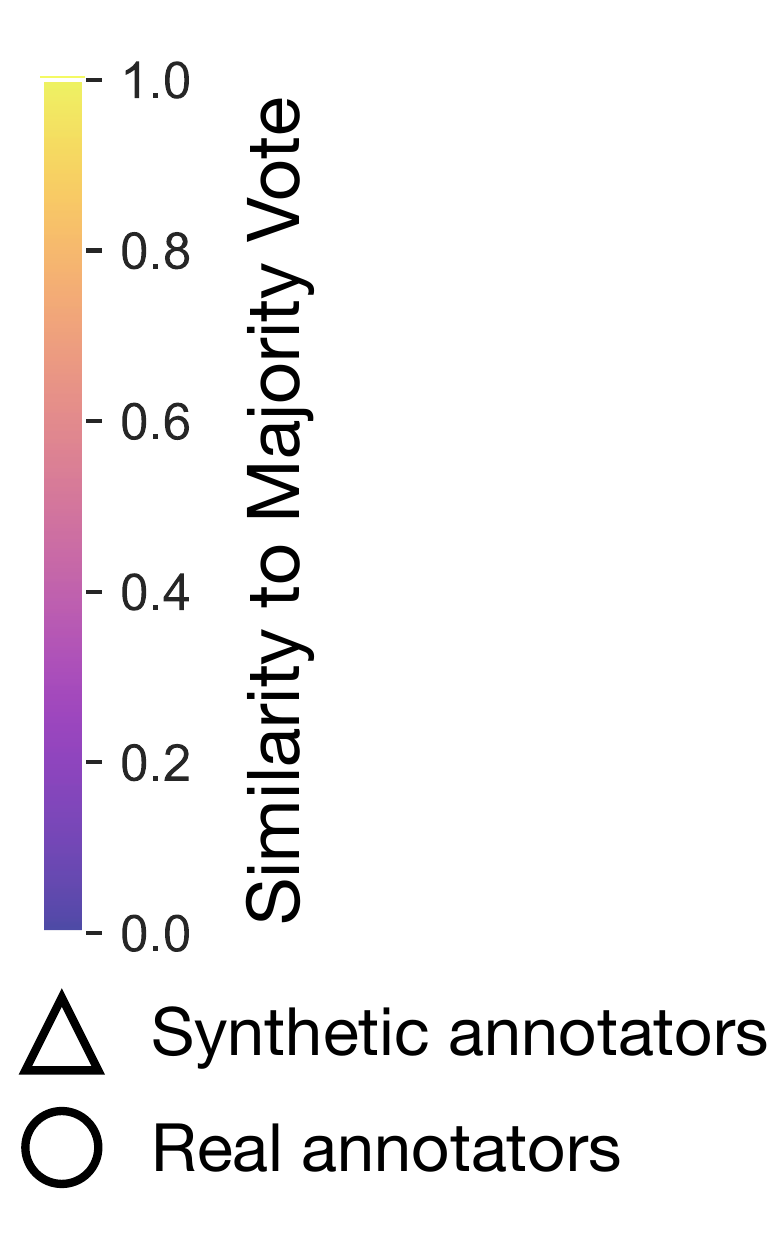}
     \end{subfigure}
    \caption{The UMAP projection visualizes the annotator embeddings, with the colorbar indicating the fraction of annotator votes aligning with the majority vote. The smaller circles show the real annotators. Larger triangles represent two distinct sets of synthetic annotators integrated into the annotator pool for each dataset. Within each group of eight synthetic annotators, the yellow triangles consistently assign the majority vote, while the purple triangles consistently diverge from the majority. A distinct separation is evident between annotators aligning more closely with the majority opinion and those deviating further from the majority vote in their annotation patterns.}
    \label{fig:umaps}
    
\end{figure*}

\subsection{Quantitative Analysis of Performance}
\label{sec:quant}
Table \ref{tab:results} presents the numerical outcomes based on the metrics outlined in \S\ref{sec:eval}. The experimental setup is described in \S\ref{sec:exp_setup}. Given that the primary motivation in this paper is to achieve equitable performance across all annotators, we initially focus on the \textit{Annotator-level F1}. Across all three datasets, one of the two variants of the AART model (with or without the use of contrastive loss) consistently outperforms both single-task and multi-task models. This underscores the AART model's capacity to excel across diverse annotators, irrespective of their contribution counts or other characteristics.

It's noteworthy that incorporating contrastive loss with AART proves particularly beneficial for the Irony detection dataset ($\mathcal{D}_{\textsl{I}}$). Without contrastive loss, the AART model faces challenges in converging on this dataset, possibly due to the inherent difficulty of automatically recognizing irony in text. The generally lower overall performances on $\mathcal{D}_{\textsl{I}}$ compared to the other two datasets affirm the challenging nature of this particular dataset. Meanwhile, the disparities in AART's performance with or without the contrastive loss on the other two datasets are marginal, indicating the overall benefit of including the contrastive loss in the training objective.

The \textit{Global-level F1} metric assesses how well the predicted label for an annotator-item pair aligns with the true label. Notably, whenever the AART approach outperforms the two baselines on \textit{Annotator-level F1}, it also excels on \textit{Global-level F1}. This observation indicates that AART achieves a robust \textit{Annotator-level F1} without merely overfitting to sparse annotators, maintaining a comprehensive understanding of item-annotator relationships across all pairs in the dataset.

The Pearson correlations between item-level labels disagreements, derived from true labels versus predicted labels, are presented in the third section of Table \ref{tab:results}. The capacity to capture disagreements among annotators at the item level is a major advantage of multi-annotator models over conventional single-task models. The results demonstrate the effectiveness of utilizing contrastive loss ($\alpha>0$) in improving the correlation of item-level disagreements when compared to the \textit{AART} model, which does not employ contrastive loss ($\alpha=0$). This enhancement is observed in the case of both $\mathcal{D}_{\textsl{I}}$ and $\mathcal{D}_{\textsl{RB}}$ datasets. However, in the case of $\mathcal{D}_{\textsl{MDA}}$, there is a marginal decline in performance, although it still outperforms the multi-task model.

\subsection{Annotation Patterns Captured in Embeddings}
In this section, we aim to gain a deeper understanding of the annotation patterns captured in the annotator embeddings. We are particularly interested in exploring whether the embeddings are influenced by the annotators' tendency to align with the majority opinion. 
To investigate this, we added two sets of eight synthetic annotators into the annotator pool for each dataset. Each synthetic annotator was exposed to $1/8$ of the data items. Within each set, eight annotators consistently provided the majority vote (determined from true labels) for the items, while the remaining eight consistently assigned a label different from the majority vote.

Upon completion of the AART training, we illustrate the embeddings in a 2D plot using UMAP projection \cite{mcinnes2018umap}. Figure \ref{fig:umaps} demonstrates that the two groups of synthetic annotators exhibit distinct separation from each other. Also, the real annotators aligning more closely with the majority opinion and those deviating further from the majority vote in their annotation patterns are showing divergence in the embeddings. This observation shows the similarity to the majority opinion indeed plays a significant role in shaping the annotator embeddings.

We encourage future research to investigate additional patterns of similarity among annotators. These similarities may stem from shared demographic features or the conditions in which annotators have been provided with prompts for the annotation task \cite{sap-etal-2019-risk}.

\subsection{Fairness to Marginalized Annotators}
\label{sec:fairness}

\begin{table}[ht!]
\centering
\small
\begin{tabular}{llll}
\hline
\multicolumn{4}{c}{Similarity to Majority}                                                          \\ \hline
       & Single-Task                & Multi-task                   & AART                         \\
$\mathcal{D}_{\textsl{MDA}}$ & $0.14 \pm 0.02$ & $0.07 \pm 0.01$ & $0.07 \pm 0.01$ \\
$\mathcal{D}_{\textsl{I}}$   & $0.14 \pm 0.02$ & $0.12 \pm 0.02$ & $0.12 \pm 0.02$ \\
$\mathcal{D}_{\textsl{RB}}$  & $0.20 \pm 0.04$ & $0.04 \pm 0.03$ & $0.05 \pm 0.03$ \\ \hline
\multicolumn{4}{c}{Count of Contributions}                                                          \\ \hline
       & Single-Task                & Multi-task                   & AART                         \\
$\mathcal{D}_{\textsl{MDA}}$ & $0.03 \pm 0.02$ & $0.14 \pm 0.04$ & $0.06 \pm 0.02$ \\
$\mathcal{D}_{\textsl{I}}$   & $0.03 \pm 0.02$ & $0.03 \pm 0.02$ & $0.02 \pm 0.02$ \\
$\mathcal{D}_{\textsl{RB}}$  & $0.07 \pm 0.03$ & $0.07 \pm 0.05$ & $0.03 \pm 0.03$ \\ \hline
\end{tabular}
\caption{Statistical Parity based on two sensitive attributes for the annotators: (1) Similarity to majority vote and (2) Count of Contributions. Mean and Standard deviations are reported based on running with 10 different random seeds.}
\label{tab:fairness}
\end{table}

In this section, we assess the model's ability to capture the perspectives of annotators who exhibit minority opinion, and those who contribute few annotations. We do this through the lens of Statistical Parity, a popular metric for algorithmic fairness~\cite{mehrabi2021survey}. Statistical parity measures the relative rate of a successful outcome for the majority group against that of the minority group. In separate experiments, we study the majority and minority groups from two perspectives: (1) the similarity of annotation selections compared to the majority vote and (2) the count of contributions. In (1), the grouping of annotators is determined by the fraction of annotations generated by a labeler that are equal to the majority vote of the items they annotated. The annotators that fall in the lowest 25th percentile in terms of similarity to the majority vote are considered in the minority group and the rest fall into the majority group. In (2), the minority group of annotators is determined by the lowest 25th percentile in terms of the number of contributions. We compare Statistical Parity across \textit{Single-task}, \textit{Multi-task}, and \textit{AART}. 
Our interest lies in showing whether each method performs equally accurate for the sensitive group versus the rest of annotators. 

Table \ref{tab:fairness} displays the results of disparity among the two groups of annotators. The lower disparity values indicates the predictions are more fair. Observing the grouping of annotators based on their similarity to majority behavior, we note that the Single-task approach which consistently predicts the majority vote exhibits the most significant disparity in accuracy between the two groups. Additionally, AART and Multi-task show similar levels of disparity, which is less compared to the Single-task approach. This observation confirms that aggregating labels causes unfairness in accuracy toward annotators whose opinions differ from the majority behavior. And motivates the use of unaggregated annotations and multi-perspective classification models instead of aggregated ground truths. 

When we group annotators based on their count of contributions, Multi-task shows the most disparity among the sensitive group of annotators and others, while AART is more equitable than the other two methods. This observation confirms that Multi-task is limited for sparse annotators and performs poorly for them, whereas AART demonstrates more balanced performance among sparse and prolific annotators.


\section{Discussion}

\begin{table*}[ht]
\centering
\small
\begin{tabular}{lllll}
\multicolumn{1}{l|}{\textit{Dataset}}       & \textit{Single-task }     & \textit{Multi-task}  & \textit{AART} ($\alpha=0$) & \textit{AART} ($\alpha>0$) \\ \hline
\multicolumn{5}{c}{\textit{Aggregated Label Performance}}                                                                                          \\ \hline
\multicolumn{1}{l|}{$\mathcal{D}_{\textsl{MDA}}$}             & 78.58±0.9            & 78.99±1.3           & 79.51±1.0          & 79.66±1.2                       \\
\multicolumn{1}{l|}{$\mathcal{D}_{\textsl{I}}$}               & 64.05±2.9            & 65.47±2.5           & 60.22±9.4          & 64.95±2.1                       \\
\multicolumn{1}{l|}{$\mathcal{D}_{\textsl{RB}}$}              & 81.09±3.5            & 81.43±3.1           & 81.43±3.5          & 81.02±1.6                      
\end{tabular}
\caption{In this table the performance (F1) of the three models on predicting the aggregated label is reported. The single-task model computes the majority labels directly. For both the multi-task model and our proposed method (AART), we determined the majority of predictions across labels for each annotator.}
\label{tab:aggr_result}
\end{table*}

Predicting a single label remains the most popular problem in classification applications. Previous studies have demonstrated that multi-annotator models can perform comparably to single-task models in predicting the aggregated labels \cite{davani-etal-2022-dealing, hayat_sensors, al-kuwatly-etal-2020-identifying}. In the single-task model, the aggregated labels are computed directly, while in multi-annotator models, the aggregated label is derived by consolidating predictions across annotators. We assess the performance of predicting the aggregated label and present the results in Table \ref{tab:aggr_result}. Our findings align with previous observations, indicating that all three models effectively predict single label outcomes.

An additional advantage of training multi-annotator models compared to single-task models is the ability to learn the entire label distribution rather than solely focusing on aggregated labels. This flexibility proves valuable in various real-world scenarios. As models are increasingly tailored for more personal, subjective use cases, a direct application of this work could be to predict personalized preferences for each annotator. 

Furthermore, \cite{gordon2022jury} introduces an inference approach for trained multi-annotator models, advocating for the selection of a task-specific subset of annotators akin to a ``Jury.'' For instance, for the task of toxicity detection in online posts, prioritizing votes from women and Black individuals---often targeted by online harassment---could prove advantageous. Their goal with jury learning is to assist practitioners in acknowledging and incorporating annotator disagreement within the classifier pipeline. To achieve this, they encourage practitioners to form a jury that specifies the individuals or groups their classifiers should emulate.

Rather than completely disregarding the diversity present in human labels, our paper proposes an effective approach to predict individual perspectives while maintaining competence in predicting the aggregated label. Instead of eliminating disagreement at the initial stages of machine learning, our aim is to learn from the various perspectives of annotators. By embracing the diversity of opinions through data-driven pipelines, which enhances data transparency, we are able to present it to end-users and human decision-makers. This empowerment allows them to take corrective actions and interventions by leveraging common sense and contextual information.

\section{Conclusion}
This paper tackles the challenges posed by annotator disagreements in subjective tasks, where establishing a single ground truth is not straightforward. We propose an approach named Annotator Aware Representations for Texts (AART) for modeling annotators' perspectives. Our methodology combines representations of individual annotators with those of text in the classification task. By learning the embeddings of annotators through Cross Entropy loss, contrastive loss, and regularization, we capture their individual labeling patterns. We compare our approach to a popular multi-task architecture, which is constrained by under-trained classification heads for annotators with limited annotations. Through several experiments, we demonstrate that AART, despite its simplicity, effectively captures the semantics of textual data for classification tasks while accommodating different annotators' perspectives. Moreover, we show that AART is fair and equitable when it comes to marginalized annotators, who may have few labeled samples or provide votes that often differ from the majority votes for samples.

\section*{Limitations}
The performance of the models outlined in this paper may be adversely affected when dealing with noisy annotations. This noise can arise from unclear instructions or insufficient attention from the worker. A limitation of this study, as well as many others focusing on designing multi-annotator models, is the tendency to overlook the high likelihood of certain annotators being unreliable. Differentiating between labeling noise and natural disagreement is a challenge that remains unaddressed in this work, and in numerous other studies.  Additionally, the degree of overlap among items annotated by multiple individuals can enhance the learning process for multi-annotator models by exposing both similarities and differences among the annotators. Nevertheless, crowdsourced datasets are inherently imperfect and susceptible to noise or sparsity. It is imperative for future research to examine the robustness of multi-annotator models in the face of sub-optimal dataset features.


\section*{Ethical Considerations}
This work seeks to improve AI systems by incorporating perspectives beyond just the most prolific annotators, as is currently done in most AI systems. This will encourage developers of subsequent AI systems to recruit more sparse annotators, as the difference in perspective they bring will advance the performance of their resulting models, rather than hinder it. This advances the ACM and ACL ethics code's core tenet that ``all people are stakeholders in computing,'' and to ``take action not to discriminate.''\footnote{https://www.acm.org/code-of-ethics} In this case, the discrimination would be against those who do not provide a wealth of annotations, due to time or other constraints.

As with any approach, there is the possibility that it could cause ethical harm. For example, a coordinated and nefarious minority could have amplified sway in our model compared to one that only focuses on the most prolific annotators. Furthermore, research has shown that large language models, such as the ones we use in this work, capture various biases that can harm our proposed model. Thus, the perspective we learn for each annotator is through the lens of these models, and it might be different from reality. Nevertheless, we maintain that this potential harm is vastly outweighed by the potential of incorporating all voices.

%
\section*{Acknowledgements}
We express gratitude to the anonymous reviewers for providing valuable feedback and offering suggestions for our project. This work was funded in part by Defense Advanced Research Projects Agency (DARPA) and Army Research Office (ARO) under Contract No. W911NF-21-C-0002 and by the {\em Multilingual Perspective-Aware NLU} project in partnership with Amazon Alexa.

\bibliography{anthology,custom}

\begin{thebibliography}{49}
\expandafter\ifx\csname natexlab\endcsname\relax\def\natexlab#1{#1}\fi

\bibitem[{Akhtar et~al.(2020)Akhtar, Basile, and Patti}]{akhtar2020modeling}
Sohail Akhtar, Valerio Basile, and Viviana Patti. 2020.
\newblock Modeling annotator perspective and polarized opinions to improve hate speech detection.
\newblock In \emph{Proceedings of the AAAI Conference on Human Computation and Crowdsourcing}, volume~8, pages 151--154.

\bibitem[{Al~Kuwatly et~al.(2020)Al~Kuwatly, Wich, and Groh}]{al-kuwatly-etal-2020-identifying}
Hala Al~Kuwatly, Maximilian Wich, and Georg Groh. 2020.
\newblock \href {https://doi.org/10.18653/v1/2020.alw-1.21} {Identifying and measuring annotator bias based on annotators{'} demographic characteristics}.
\newblock In \emph{Proceedings of the Fourth Workshop on Online Abuse and Harms}, pages 184--190, Online. Association for Computational Linguistics.

\bibitem[{Aroyo et~al.(2024)Aroyo, Taylor, Diaz, Homan, Parrish, Serapio-Garc{\'\i}a, Prabhakaran, and Wang}]{aroyo2024dices}
Lora Aroyo, Alex Taylor, Mark Diaz, Christopher Homan, Alicia Parrish, Gregory Serapio-Garc{\'\i}a, Vinodkumar Prabhakaran, and Ding Wang. 2024.
\newblock Dices dataset: Diversity in conversational ai evaluation for safety.
\newblock \emph{Advances in Neural Information Processing Systems}, 36.

\bibitem[{Aroyo and Welty(2014)}]{aroyo2014three}
Lora Aroyo and Chris Welty. 2014.
\newblock The three sides of crowdtruth.
\newblock \emph{Human Computation}, 1(1).

\bibitem[{Aroyo and Welty(2015)}]{Aroyo_Welty_2015}
Lora Aroyo and Chris Welty. 2015.
\newblock \href {https://doi.org/10.1609/aimag.v36i1.2564} {Truth is a lie: Crowd truth and the seven myths of human annotation}.
\newblock \emph{AI Magazine}, 36(1):15--24.

\bibitem[{Basile(2021)}]{basile2021s}
Valerio Basile. 2021.
\newblock It’s the end of the gold standard as we know it: Leveraging non-aggregated data for better evaluation and explanation of subjective tasks.
\newblock In \emph{AIxIA 2020--Advances in Artificial Intelligence: XIXth International Conference of the Italian Association for Artificial Intelligence, Virtual Event, November 25--27, 2020, Revised Selected Papers}, pages 441--453. Springer.

\bibitem[{Basile et~al.(2021{\natexlab{a}})Basile, Cabitza, Campagner, and Fell}]{basile2021toward}
Valerio Basile, Federico Cabitza, Andrea Campagner, and Michael Fell. 2021{\natexlab{a}}.
\newblock Toward a perspectivist turn in ground truthing for predictive computing.
\newblock \emph{arXiv preprint arXiv:2109.04270}.

\bibitem[{Basile et~al.(2021{\natexlab{b}})Basile, Fell, Fornaciari, Hovy, Paun, Plank, Poesio, and Uma}]{basile-etal-2021-need}
Valerio Basile, Michael Fell, Tommaso Fornaciari, Dirk Hovy, Silviu Paun, Barbara Plank, Massimo Poesio, and Alexandra Uma. 2021{\natexlab{b}}.
\newblock \href {https://doi.org/10.18653/v1/2021.bppf-1.3} {We need to consider disagreement in evaluation}.
\newblock In \emph{Proceedings of the 1st Workshop on Benchmarking: Past, Present and Future}, pages 15--21, Online. Association for Computational Linguistics.

\bibitem[{Chen et~al.(2020)Chen, Kornblith, Norouzi, and Hinton}]{chen2020simple}
Ting Chen, Simon Kornblith, Mohammad Norouzi, and Geoffrey Hinton. 2020.
\newblock A simple framework for contrastive learning of visual representations.
\newblock In \emph{International conference on machine learning}, pages 1597--1607. PMLR.

\bibitem[{Deng et~al.(2023)Deng, Zhang, Liu, Wu, Wang, and Mihalcea}]{deng-etal-2023-annotate}
Naihao Deng, Xinliang Zhang, Siyang Liu, Winston Wu, Lu~Wang, and Rada Mihalcea. 2023.
\newblock \href {https://doi.org/10.18653/v1/2023.findings-emnlp.832} {You are what you annotate: Towards better models through annotator representations}.
\newblock In \emph{Findings of the Association for Computational Linguistics: EMNLP 2023}, pages 12475--12498, Singapore. Association for Computational Linguistics.

\bibitem[{Denton et~al.(2021)Denton, D{\'\i}az, Kivlichan, Prabhakaran, and Rosen}]{denton2021whose}
Emily Denton, Mark D{\'\i}az, Ian Kivlichan, Vinodkumar Prabhakaran, and Rachel Rosen. 2021.
\newblock Whose ground truth? accounting for individual and collective identities underlying dataset annotation.
\newblock \emph{arXiv preprint arXiv:2112.04554}.

\bibitem[{Devlin et~al.(2019)Devlin, Chang, Lee, and Toutanova}]{devlin-etal-2019-bert}
Jacob Devlin, Ming-Wei Chang, Kenton Lee, and Kristina Toutanova. 2019.
\newblock \href {https://doi.org/10.18653/v1/N19-1423} {{BERT}: Pre-training of deep bidirectional transformers for language understanding}.
\newblock In \emph{Proceedings of the 2019 Conference of the North {A}merican Chapter of the Association for Computational Linguistics: Human Language Technologies, Volume 1 (Long and Short Papers)}, pages 4171--4186, Minneapolis, Minnesota. Association for Computational Linguistics.

\bibitem[{Dumitrache et~al.(2019)Dumitrache, Aroyo, and Welty}]{dumitrache-etal-2019-crowdsourced}
Anca Dumitrache, Lora Aroyo, and Chris Welty. 2019.
\newblock \href {https://doi.org/10.18653/v1/N19-1224} {A crowdsourced frame disambiguation corpus with ambiguity}.
\newblock In \emph{Proceedings of the 2019 Conference of the North {A}merican Chapter of the Association for Computational Linguistics: Human Language Technologies, Volume 1 (Long and Short Papers)}, pages 2164--2170, Minneapolis, Minnesota. Association for Computational Linguistics.

\bibitem[{Dumitrache et~al.(2018)Dumitrache, Inel, Aroyo, Timmermans, and Welty}]{dumitrache2018crowdtruth}
Anca Dumitrache, Oana Inel, Lora Aroyo, Benjamin Timmermans, and Chris Welty. 2018.
\newblock Crowdtruth 2.0: Quality metrics for crowdsourcing with disagreement.
\newblock \emph{arXiv preprint arXiv:1808.06080}.

\bibitem[{Fornaciari et~al.(2021)Fornaciari, Uma, Paun, Plank, Hovy, and Poesio}]{fornaciari-etal-2021-beyond}
Tommaso Fornaciari, Alexandra Uma, Silviu Paun, Barbara Plank, Dirk Hovy, and Massimo Poesio. 2021.
\newblock \href {https://doi.org/10.18653/v1/2021.naacl-main.204} {Beyond black {\&} white: Leveraging annotator disagreement via soft-label multi-task learning}.
\newblock In \emph{Proceedings of the 2021 Conference of the North American Chapter of the Association for Computational Linguistics: Human Language Technologies}, pages 2591--2597, Online. Association for Computational Linguistics.

\bibitem[{Frenda et~al.(2023)Frenda, Pedrani, Basile, Lo, Cignarella, Panizzon, Marco, Scarlini, Patti, Bosco, and Bernardi}]{frenda-etal-2023-epic}
Simona Frenda, Alessandro Pedrani, Valerio Basile, Soda~Marem Lo, Alessandra~Teresa Cignarella, Raffaella Panizzon, Cristina Marco, Bianca Scarlini, Viviana Patti, Cristina Bosco, and Davide Bernardi. 2023.
\newblock \href {https://doi.org/10.18653/v1/2023.acl-long.774} {{EPIC}: Multi-perspective annotation of a corpus of irony}.
\newblock In \emph{Proceedings of the 61st Annual Meeting of the Association for Computational Linguistics (Volume 1: Long Papers)}, pages 13844--13857, Toronto, Canada. Association for Computational Linguistics.

\bibitem[{Gordon et~al.(2022)Gordon, Lam, Park, Patel, Hancock, Hashimoto, and Bernstein}]{gordon2022jury}
Mitchell~L Gordon, Michelle~S Lam, Joon~Sung Park, Kayur Patel, Jeff Hancock, Tatsunori Hashimoto, and Michael~S Bernstein. 2022.
\newblock Jury learning: Integrating dissenting voices into machine learning models.
\newblock In \emph{Proceedings of the 2022 CHI Conference on Human Factors in Computing Systems}, pages 1--19.

\bibitem[{Gordon et~al.(2021)Gordon, Zhou, Patel, Hashimoto, and Bernstein}]{gordon2021disagreement}
Mitchell~L Gordon, Kaitlyn Zhou, Kayur Patel, Tatsunori Hashimoto, and Michael~S Bernstein. 2021.
\newblock The disagreement deconvolution: Bringing machine learning performance metrics in line with reality.
\newblock In \emph{Proceedings of the 2021 CHI Conference on Human Factors in Computing Systems}, pages 1--14.

\bibitem[{Hara et~al.(2019)Hara, Adams, Milland, Savage, Hanrahan, Bigham, and Callison-Burch}]{hara2019worker}
Kotaro Hara, Abigail Adams, Kristy Milland, Saiph Savage, Benjamin~V Hanrahan, Jeffrey~P Bigham, and Chris Callison-Burch. 2019.
\newblock Worker demographics and earnings on amazon mechanical turk: An exploratory analysis.
\newblock In \emph{Extended abstracts of the 2019 chi conference on human factors in computing systems}, pages 1--6.

\bibitem[{Hayat et~al.(2022)Hayat, Ventura, and Lapedriza}]{hayat_sensors}
Hassan Hayat, Carles Ventura, and Agata Lapedriza. 2022.
\newblock \href {https://doi.org/10.3390/s22145245} {Modeling subjective affect annotations with multi-task learning}.
\newblock \emph{Sensors}, 22(14).

\bibitem[{He et~al.(2021)He, Tavabi, Lerman, and Soleymani}]{he-etal-2021-speaker-turn}
Zihao He, Leili Tavabi, Kristina Lerman, and Mohammad Soleymani. 2021.
\newblock \href {https://doi.org/10.18653/v1/2021.findings-emnlp.185} {Speaker turn modeling for dialogue act classification}.
\newblock In \emph{Findings of the Association for Computational Linguistics: EMNLP 2021}, pages 2150--2157, Punta Cana, Dominican Republic. Association for Computational Linguistics.

\bibitem[{Hovy et~al.(2013)Hovy, Berg-Kirkpatrick, Vaswani, and Hovy}]{hovy-etal-2013-learning}
Dirk Hovy, Taylor Berg-Kirkpatrick, Ashish Vaswani, and Eduard Hovy. 2013.
\newblock \href {https://aclanthology.org/N13-1132} {Learning whom to trust with {MACE}}.
\newblock In \emph{Proceedings of the 2013 Conference of the North {A}merican Chapter of the Association for Computational Linguistics: Human Language Technologies}, pages 1120--1130, Atlanta, Georgia. Association for Computational Linguistics.

\bibitem[{Kenyon-Dean et~al.(2018)Kenyon-Dean, Ahmed, Fujimoto, Georges-Filteau, Glasz, Kaur, Lalande, Bhanderi, Belfer, Kanagasabai, Sarrazingendron, Verma, and Ruths}]{kenyon-dean-etal-2018-sentiment}
Kian Kenyon-Dean, Eisha Ahmed, Scott Fujimoto, Jeremy Georges-Filteau, Christopher Glasz, Barleen Kaur, Auguste Lalande, Shruti Bhanderi, Robert Belfer, Nirmal Kanagasabai, Roman Sarrazingendron, Rohit Verma, and Derek Ruths. 2018.
\newblock \href {https://doi.org/10.18653/v1/N18-1171} {Sentiment analysis: It{'}s complicated!}
\newblock In \emph{Proceedings of the 2018 Conference of the North {A}merican Chapter of the Association for Computational Linguistics: Human Language Technologies, Volume 1 (Long Papers)}, pages 1886--1895, New Orleans, Louisiana. Association for Computational Linguistics.

\bibitem[{Kingma and Ba(2014)}]{kingma2014adam}
Diederik~P Kingma and Jimmy Ba. 2014.
\newblock Adam: A method for stochastic optimization.
\newblock \emph{arXiv preprint arXiv:1412.6980}.

\bibitem[{Larimore et~al.(2021)Larimore, Kennedy, Haskett, and Arseniev-Koehler}]{larimore-etal-2021-reconsidering}
Savannah Larimore, Ian Kennedy, Breon Haskett, and Alina Arseniev-Koehler. 2021.
\newblock \href {https://doi.org/10.18653/v1/2021.socialnlp-1.7} {Reconsidering annotator disagreement about racist language: Noise or signal?}
\newblock In \emph{Proceedings of the Ninth International Workshop on Natural Language Processing for Social Media}, pages 81--90, Online. Association for Computational Linguistics.

\bibitem[{Leonardelli et~al.(2021)Leonardelli, Menini, Palmero~Aprosio, Guerini, and Tonelli}]{leonardelli-etal-2021-agreeing}
Elisa Leonardelli, Stefano Menini, Alessio Palmero~Aprosio, Marco Guerini, and Sara Tonelli. 2021.
\newblock \href {https://doi.org/10.18653/v1/2021.emnlp-main.822} {Agreeing to disagree: Annotating offensive language datasets with annotators{'} disagreement}.
\newblock In \emph{Proceedings of the 2021 Conference on Empirical Methods in Natural Language Processing}, pages 10528--10539, Online and Punta Cana, Dominican Republic. Association for Computational Linguistics.

\bibitem[{Liu et~al.(2019)Liu, Ott, Goyal, Du, Joshi, Chen, Levy, Lewis, Zettlemoyer, and Stoyanov}]{liu2019roberta}
Yinhan Liu, Myle Ott, Naman Goyal, Jingfei Du, Mandar Joshi, Danqi Chen, Omer Levy, Mike Lewis, Luke Zettlemoyer, and Veselin Stoyanov. 2019.
\newblock Roberta: A robustly optimized bert pretraining approach.
\newblock \emph{arXiv preprint arXiv:1907.11692}.

\bibitem[{Lo and Basile(2023)}]{DBLP:conf/nlperspectives/LoB23}
Soda~Marem Lo and Valerio Basile. 2023.
\newblock \href {https://ceur-ws.org/Vol-3494/paper8.pdf} {Hierarchical clustering of label-based annotator representations for mining perspectives}.
\newblock In \emph{Proceedings of the 2nd Workshop on Perspectivist Approaches to {NLP} co-located with 26th European Conference on Artificial Intelligence {(ECAI} 2023), Krak{\'{o}}w, Poland, September 30th, 2023}, volume 3494 of \emph{{CEUR} Workshop Proceedings}. CEUR-WS.org.

\bibitem[{McInnes et~al.(2018)McInnes, Healy, and Melville}]{mcinnes2018umap}
Leland McInnes, John Healy, and James Melville. 2018.
\newblock Umap: Uniform manifold approximation and projection for dimension reduction.
\newblock \emph{arXiv preprint arXiv:1802.03426}.

\bibitem[{Mehrabi et~al.(2021)Mehrabi, Morstatter, Saxena, Lerman, and Galstyan}]{mehrabi2021survey}
Ninareh Mehrabi, Fred Morstatter, Nripsuta Saxena, Kristina Lerman, and Aram Galstyan. 2021.
\newblock A survey on bias and fairness in machine learning.
\newblock \emph{ACM computing surveys (CSUR)}, 54(6):1--35.

\bibitem[{Mokhberian et~al.(2022)Mokhberian, Hopp, Harandizadeh, Morstatter, and Lerman}]{mokhberian2022noise}
Negar Mokhberian, Frederic~R Hopp, Bahareh Harandizadeh, Fred Morstatter, and Kristina Lerman. 2022.
\newblock Noise audits improve moral foundation classification.
\newblock In \emph{Proceedings of the 2023 IEEE/ACM International Conference on Advances in Social Networks Analysis and Mining (ASONAM)}.

\bibitem[{Mostafazadeh~Davani et~al.(2022)Mostafazadeh~Davani, D{\'\i}az, and Prabhakaran}]{davani-etal-2022-dealing}
Aida Mostafazadeh~Davani, Mark D{\'\i}az, and Vinodkumar Prabhakaran. 2022.
\newblock \href {https://doi.org/10.1162/tacl_a_00449} {Dealing with disagreements: Looking beyond the majority vote in subjective annotations}.
\newblock \emph{Transactions of the Association for Computational Linguistics}, 10:92--110.

\bibitem[{Oord et~al.(2018)Oord, Li, and Vinyals}]{oord2018representation}
Aaron van~den Oord, Yazhe Li, and Oriol Vinyals. 2018.
\newblock Representation learning with contrastive predictive coding.
\newblock \emph{arXiv preprint arXiv:1807.03748}.

\bibitem[{Orlikowski et~al.(2023)Orlikowski, R{\"o}ttger, Cimiano, and Hovy}]{orlikowski-etal-2023-ecological}
Matthias Orlikowski, Paul R{\"o}ttger, Philipp Cimiano, and Dirk Hovy. 2023.
\newblock \href {https://doi.org/10.18653/v1/2023.acl-short.88} {The ecological fallacy in annotation: Modeling human label variation goes beyond sociodemographics}.
\newblock In \emph{Proceedings of the 61st Annual Meeting of the Association for Computational Linguistics (Volume 2: Short Papers)}, pages 1017--1029, Toronto, Canada. Association for Computational Linguistics.

\bibitem[{Plank(2022)}]{plank-2022-problem}
Barbara Plank. 2022.
\newblock \href {https://doi.org/10.18653/v1/2022.emnlp-main.731} {The {``}problem{''} of human label variation: On ground truth in data, modeling and evaluation}.
\newblock In \emph{Proceedings of the 2022 Conference on Empirical Methods in Natural Language Processing}, pages 10671--10682, Abu Dhabi, United Arab Emirates. Association for Computational Linguistics.

\bibitem[{Plepi et~al.(2022)Plepi, Neuendorf, Flek, and Welch}]{plepi2022unifying}
Joan Plepi, B{\'e}la Neuendorf, Lucie Flek, and Charles Welch. 2022.
\newblock Unifying data perspectivism and personalization: An application to social norms.
\newblock \emph{EMNLP}.

\bibitem[{Posch et~al.(2022)Posch, Bleier, Flöck, Lechner, Kinder-Kurlanda, Helic, and Strohmaier}]{Posch_2022}
Lisa Posch, Arnim Bleier, Fabian Flöck, Clemens~M. Lechner, Katharina Kinder-Kurlanda, Denis Helic, and Markus Strohmaier. 2022.
\newblock \href {https://doi.org/10.15346/hc.v9i1.106} {Characterizing the global crowd workforce: A cross-country comparison of crowdworker demographics}.
\newblock \emph{Human Computation}, 9(1):22--57.

\bibitem[{Prabhakaran et~al.(2021)Prabhakaran, Mostafazadeh~Davani, and Diaz}]{prabhakaran-etal-2021-releasing}
Vinodkumar Prabhakaran, Aida Mostafazadeh~Davani, and Mark Diaz. 2021.
\newblock \href {https://doi.org/10.18653/v1/2021.law-1.14} {On releasing annotator-level labels and information in datasets}.
\newblock In \emph{Proceedings of the Joint 15th Linguistic Annotation Workshop (LAW) and 3rd Designing Meaning Representations (DMR) Workshop}, pages 133--138, Punta Cana, Dominican Republic. Association for Computational Linguistics.

\bibitem[{Rodrigues and Pereira(2018)}]{learncrowd}
Filipe Rodrigues and Francisco~C. Pereira. 2018.
\newblock Deep learning from crowds.
\newblock In \emph{Proceedings of the Thirty-Second AAAI Conference on Artificial Intelligence and Thirtieth Innovative Applications of Artificial Intelligence Conference and Eighth AAAI Symposium on Educational Advances in Artificial Intelligence}, AAAI'18/IAAI'18/EAAI'18. AAAI Press.

\bibitem[{Sap et~al.(2019)Sap, Card, Gabriel, Choi, and Smith}]{sap-etal-2019-risk}
Maarten Sap, Dallas Card, Saadia Gabriel, Yejin Choi, and Noah~A. Smith. 2019.
\newblock \href {https://doi.org/10.18653/v1/P19-1163} {The risk of racial bias in hate speech detection}.
\newblock In \emph{Proceedings of the 57th Annual Meeting of the Association for Computational Linguistics}, pages 1668--1678, Florence, Italy. Association for Computational Linguistics.

\bibitem[{Sap et~al.(2022)Sap, Swayamdipta, Vianna, Zhou, Choi, and Smith}]{sap-etal-2022-annotators}
Maarten Sap, Swabha Swayamdipta, Laura Vianna, Xuhui Zhou, Yejin Choi, and Noah~A. Smith. 2022.
\newblock \href {https://doi.org/10.18653/v1/2022.naacl-main.431} {Annotators with attitudes: How annotator beliefs and identities bias toxic language detection}.
\newblock In \emph{Proceedings of the 2022 Conference of the North American Chapter of the Association for Computational Linguistics: Human Language Technologies}, pages 5884--5906, Seattle, United States. Association for Computational Linguistics.

\bibitem[{Uma et~al.(2022)Uma, Almanea, and Poesio}]{uma2022scaling}
Alexandra Uma, Dina Almanea, and Massimo Poesio. 2022.
\newblock Scaling and disagreements: Bias, noise, and ambiguity.
\newblock \emph{Frontiers in Artificial Intelligence}, 5.

\bibitem[{Uma et~al.(2021)Uma, Fornaciari, Dumitrache, Miller, Chamberlain, Plank, Simpson, and Poesio}]{uma-etal-2021-semeval}
Alexandra Uma, Tommaso Fornaciari, Anca Dumitrache, Tristan Miller, Jon Chamberlain, Barbara Plank, Edwin Simpson, and Massimo Poesio. 2021.
\newblock \href {https://doi.org/10.18653/v1/2021.semeval-1.41} {{S}em{E}val-2021 task 12: Learning with disagreements}.
\newblock In \emph{Proceedings of the 15th International Workshop on Semantic Evaluation (SemEval-2021)}, pages 338--347, Online. Association for Computational Linguistics.

\bibitem[{Vaswani et~al.(2017)Vaswani, Shazeer, Parmar, Uszkoreit, Jones, Gomez, Kaiser, and Polosukhin}]{vaswani2017attention}
Ashish Vaswani, Noam Shazeer, Niki Parmar, Jakob Uszkoreit, Llion Jones, Aidan~N Gomez, {\L}ukasz Kaiser, and Illia Polosukhin. 2017.
\newblock Attention is all you need.
\newblock \emph{Advances in neural information processing systems}, 30.

\bibitem[{Wan et~al.(2023)Wan, Kim, and Kang}]{wan2023everyone}
Ruyuan Wan, Jaehyung Kim, and Dongyeop Kang. 2023.
\newblock \href {https://doi.org/10.1609/aaai.v37i12.26698} {Everyone’s voice matters: Quantifying annotation disagreement using demographic information}.
\newblock \emph{Proceedings of the AAAI Conference on Artificial Intelligence}, 37(12):14523--14530.

\bibitem[{Waseem(2016)}]{waseem2016you}
Zeerak Waseem. 2016.
\newblock Are you a racist or am i seeing things? annotator influence on hate speech detection on twitter.
\newblock In \emph{Proceedings of the first workshop on NLP and computational social science}, pages 138--142.

\bibitem[{Weerasooriya et~al.(2023)Weerasooriya, Ororbia, Bhensadadia, KhudaBukhsh, and Homan}]{weerasooriya-etal-2023-disagreement}
Tharindu~Cyril Weerasooriya, Alexander Ororbia, Raj Bhensadadia, Ashiqur KhudaBukhsh, and Christopher Homan. 2023.
\newblock \href {https://doi.org/10.18653/v1/2023.findings-acl.287} {Disagreement matters: Preserving label diversity by jointly modeling item and annotator label distributions with {D}is{C}o}.
\newblock In \emph{Findings of the Association for Computational Linguistics: ACL 2023}, pages 4679--4695, Toronto, Canada. Association for Computational Linguistics.

\bibitem[{Wich et~al.(2020)Wich, Al~Kuwatly, and Groh}]{wich-etal-2020-investigating}
Maximilian Wich, Hala Al~Kuwatly, and Georg Groh. 2020.
\newblock \href {https://doi.org/10.18653/v1/2020.alw-1.22} {Investigating annotator bias with a graph-based approach}.
\newblock In \emph{Proceedings of the Fourth Workshop on Online Abuse and Harms}, pages 191--199, Online. Association for Computational Linguistics.

\bibitem[{Wolf et~al.(2020)Wolf, Debut, Sanh, Chaumond, Delangue, Moi, Cistac, Rault, Louf, Funtowicz, Davison, Shleifer, von Platen, Ma, Jernite, Plu, Xu, Le~Scao, Gugger, Drame, Lhoest, and Rush}]{wolf-etal-2020-transformers}
Thomas Wolf, Lysandre Debut, Victor Sanh, Julien Chaumond, Clement Delangue, Anthony Moi, Pierric Cistac, Tim Rault, Remi Louf, Morgan Funtowicz, Joe Davison, Sam Shleifer, Patrick von Platen, Clara Ma, Yacine Jernite, Julien Plu, Canwen Xu, Teven Le~Scao, Sylvain Gugger, Mariama Drame, Quentin Lhoest, and Alexander Rush. 2020.
\newblock \href {https://doi.org/10.18653/v1/2020.emnlp-demos.6} {Transformers: State-of-the-art natural language processing}.
\newblock In \emph{Proceedings of the 2020 Conference on Empirical Methods in Natural Language Processing: System Demonstrations}, pages 38--45, Online. Association for Computational Linguistics.

\end{thebibliography}
\bibliographystyle{acl_natbib}

\appendix

\section{Experimental Setup}
\label{sec:exp_setup}
\paragraph{Common parameters:}
We conducted our experiments using a shared server in our institution with a GPU NVIDIA GeForce RTX 2080 Ti 11GB. Across all three approaches we set max sequence length to 100, training batch size to 100, and learning rate to $5\mathrm{e}^{-5}$. The training steps use the Adam optimizer \cite{kingma2014adam} with a weight decay rate equal to 0.01.
We use early stopping with 20 max number of epochs and we save the model parameters from the epoch with best annotator-level F1.
For the language model we use ``roberta-base" for $\mathcal{D}_{\textsl{MDA}}$ and $\mathcal{D}_{\textsl{I}}$ datasets. However, because the $\mathcal{D}_{\textsl{RB}}$ is a much smaller dataset (only contains 1309 text items) we use a language model pretrained on the task of offensiveness detection ``cardiffnlp/twitter-roberta-base-offensive"\footnote{\url{https://huggingface.co/cardiffnlp/twitter-roberta-base-offensive}}. We load the transformer models from HuggingFace Library \cite{wolf-etal-2020-transformers}. 

We initiate the splitting process with a train/dev/test split of 50/25/25 percent stratified on item-level disagreements, after which we transfer the items from dev and test splits annotated by an annotator which is unseen in the training set to augment the training data. We repeat the splitting with 10 random seeds. 

\paragraph{AART parameters:}
We set $\tau=0.07$ for contrastive loss and performed hyper-parameter search for $\alpha \in \{0.1, 0.2, 0.3\}$ and for $\lambda \in \{0.0, 0.1, 0.2\}$ which are respectively coefficients for contrastive loss and for $L_2$ norm regularization in the objective function.

\end{document}